\documentclass[conference]{IEEEtran}
\IEEEoverridecommandlockouts

\usepackage{paper}
\usepackage[utf8]{inputenc}
\usepackage{newunicodechar}
\usepackage{balance}

\newunicodechar{ }{\,}
\def\BibTeX{{\rm B\kern-.05em{\sc i\kern-.025em b}\kern-.08em
    T\kern-.1667em\lower.7ex\hbox{E}\kern-.125emX}}

\begin{document}

\title{Reasoning LLMs in the Medical Domain: \\ A Literature Survey}

\author{
Armin Berger\IEEEauthorrefmark{9}\IEEEauthorrefmark{2}\IEEEauthorrefmark{3}\IEEEauthorrefmark{1}, \IEEEauthorblockN{Sarthak Khanna\IEEEauthorrefmark{2}\IEEEauthorrefmark{1},
David Berghaus\IEEEauthorrefmark{9}, Rafet Sifa\IEEEauthorrefmark{9}\IEEEauthorrefmark{2}}
\IEEEauthorblockA{\IEEEauthorrefmark{9}Fraunhofer IAIS - Department of Media Engineering, Germany}
\IEEEauthorblockA{\IEEEauthorrefmark{2}University of Bonn - Department of Computer Science, Germany}
\IEEEauthorblockA{\IEEEauthorrefmark{3}West-AI - Federal Ministry of Education and Research, Germany}

\thanks{* Both authors contributed equally to this research.}\\
\thanks{The project was funded by the Federal Ministry of Education and Research (BMBF) under grant no. 01IS22094A WEST-AI. This research has been partially funded by the Federal Ministry of Education and Research of Germany and the state of North-Rhine Westphalia as part of the Lamarr-Institute for Machine Learning and Artificial Intelligence.}
}

% \author{
% \IEEEauthorblockN{Anonymous}
% }

\maketitle

\begin{abstract}
% The emergence of advanced reasoning capabilities in Large Language Models (LLMs) represents a transformative development for healthcare applications. This survey examines how medical LLMs have evolved from basic information retrieval systems to sophisticated reasoning architectures supporting complex clinical decision-making. We analyze key enabling technologies, including specialized Chain-of-Thought prompting and recent Reinforcement Learning breakthroughs exemplified by DeepSeek-R1. The survey evaluates purpose-built medical frameworks such as Med-PaLM and DiagnosisGPT, alongside multi-agent collaborative systems and innovative prompting strategies. We critically assess evaluation methodologies for medical validation and address persistent challenges including interpretability, bias mitigation, patient safety, and multimodal data integration. This synthesis provides a roadmap for developing trustworthy LLMs as partners in clinical practice and medical research to enhance healthcare outcomes.
The emergence of advanced reasoning capabilities in Large Language Models (LLMs) marks a transformative development in healthcare applications. Beyond merely expanding functional capabilities, these reasoning mechanisms enhance decision transparency and explainability-critical requirements in medical contexts. This survey examines the transformation of medical LLMs from basic information retrieval tools to sophisticated clinical reasoning systems capable of supporting complex healthcare decisions. We provide a thorough analysis of the enabling technological foundations, with a particular focus on specialized prompting techniques like Chain-of-Thought and recent breakthroughs in Reinforcement Learning exemplified by DeepSeek-R1. Our investigation evaluates purpose-built medical frameworks while also examining emerging paradigms such as multi-agent collaborative systems and innovative prompting architectures. The survey critically assesses current evaluation methodologies for medical validation and addresses persistent challenges in field interpretation limitations, bias mitigation strategies, patient safety frameworks, and integration of multimodal clinical data. Through this survey, we seek to establish a roadmap for developing reliable LLMs that can serve as effective partners in clinical practice and medical research.

\end{abstract}


\section{Introduction}
The integration of Artificial Intelligence (AI) into healthcare has promised to revolutionize medical practice, from diagnostics to personalized medicine \cite{thirunavukarasu2023large}. Among AI's most dynamic subfields, Large Language Models (LLMs) have recently demonstrated remarkable capabilities in understanding, generating, and manipulating human language, leading to their exploration in numerous specialized domains \cite{moor2023foundation}. Within medicine, the demand for systems that can not only process vast amounts of information but also engage in complex reasoning is particularly acute \cite{yang2023large}. Thus, the focus has intensified on "reasoning LLMs," which are commonly defined as models that tackle problems through complex, multi-step generation with intermediate steps and often manifest a thought or thinking process in their outputs \cite{Lietal}.

This survey provides a comprehensive examination of the current state and future potential of reasoning LLMs within the medical domain. These advanced models are poised to transform critical healthcare functions, including diagnostic processes, treatment planning, drug discovery, and clinical decision support systems \cite{zheng2025large}. We explore the foundational underpinnings of medical reasoning in LLMs, chart the evolution of key techniques such as Chain-of-Thought prompting, and critically examine the emerging role of Reinforcement Learning (RL) in cultivating deeper and more generalizable reasoning abilities, as exemplified by models like DeepSeek-R1 in general AI \cite{guo2025deepseek} and Med-R1 in medical vision-language understanding \cite{lai2025med}. We discuss prominent specialized medical LLMs, multi-agent systems, and analyze the ecosystem of evaluation and optimization that supports their development. Furthermore, we critically assess the persistent challenges and delineate promising avenues for future research, aiming to foster LLMs that are not only intelligent but also safe, reliable, and ethically aligned for medical applications.

\begin{figure*}[ht]
\centering
\includegraphics[width=\textwidth]{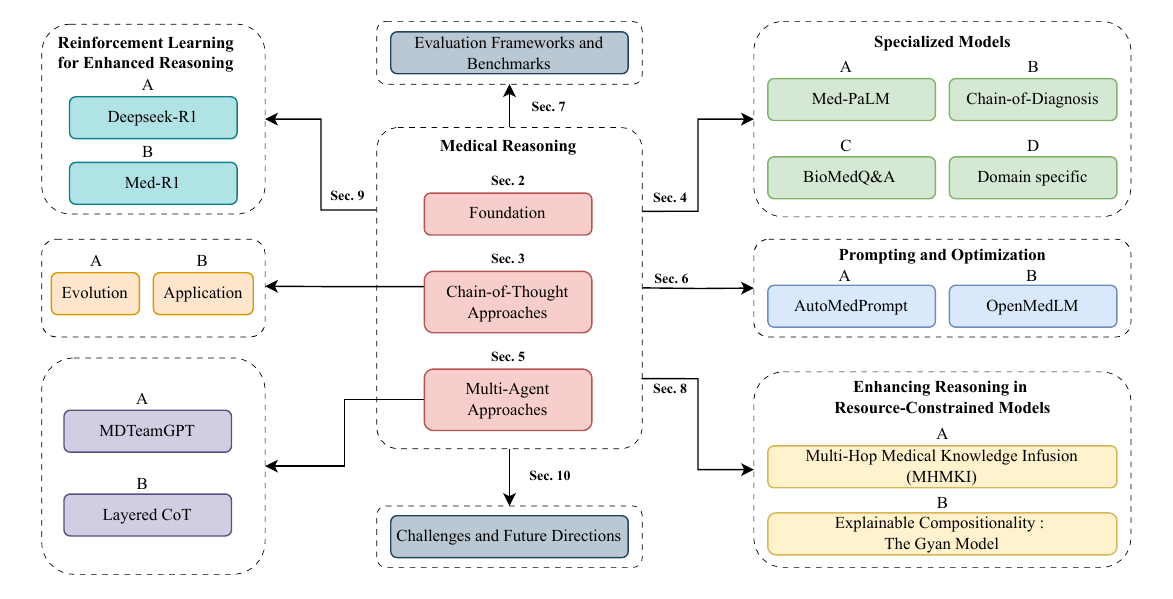}
\caption{\textbf{The schematic structure of Reasoning LLMs in the Medical Domain.}}
\label{fig:embed_pipeline}
\end{figure*}

\section{Foundations of Medical Reasoning in LLMs}

The journey towards reasoning LLMs in medicine signifies a paradigm shift from earlier AI models that primarily focused on information retrieval or narrow predictive tasks without explicit step-by-step problem-solving. Raschka's conceptualization emphasizes that a true reasoning model distinguishes itself by its proficiency in complex tasks that intrinsically benefit from decomposing the problem into intermediate, manageable steps \cite{Raschka2025Understanding}. This is a departure from models designed for simpler functions like direct factual recall or single-step classification.

Medical reasoning, however, presents a unique and formidable set of challenges compared to general-domain reasoning tasks such as solving puzzles or mathematical problems. The medical field is characterized by:
\begin{itemize}
    \item \textbf{Domain-Specific Knowledge Complexity:} Requiring deep, nuanced understanding of intricate biological systems, pathophysiology, pharmacology, and an ever-expanding corpus of medical literature.
    \item \textbf{Uncertainty and Incompleteness:} Clinical data is often noisy, incomplete, or ambiguous. Reasoning systems must manage this uncertainty, akin to how clinicians operate with probabilities and differential diagnoses.
    \item \textbf{Patient-Specificity:} Medical decisions are highly individualized, needing to account for unique patient histories, genetic predispositions, comorbidities, and socio-economic factors.
    \item \textbf{High Stakes and Safety Criticality:} Errors in medical reasoning can have severe, life-altering consequences, demanding exceptional accuracy, reliability, and safety.
\end{itemize}
Early medical LLMs, while proficient at tasks like answering medical questions from a knowledge base (e.g., BioBERT \cite{lee2020biobert}, PubMedBERT \cite{han2021pubmedbert}), often struggled with the sophisticated, multi-faceted diagnostic reasoning routinely performed by experienced physicians \cite{medpalm1}. These models could retrieve facts but lacked the deeper inferential capabilities to synthesize information into a coherent diagnostic pathway.

The inflection point towards enhanced reasoning capabilities was notably marked by the introduction of models like Med-PaLM in late 2022 \cite{medpalm1}. Med-PaLM adapted general-purpose LLMs specifically for medical question-answering, setting early benchmarks by achieving promising, though not yet expert-level, performance on standardized multiple-choice medical exams. This pioneering work illuminated a clear trajectory for subsequent research, which has increasingly concentrated on refining and augmenting the intrinsic reasoning processes of LLMs tailored for the medical domain. The emphasis shifted from mere knowledge encoding to enabling models to "think through" medical problems methodically.

\section{Chain-of-Thought Approaches in Medical AI}

Chain-of-Thought (CoT) prompting has rapidly emerged as a cornerstone technique for unlocking and guiding reasoning capabilities in LLMs, including those applied to medicine \cite{wei2022chain}. This approach prompts the model to articulate a sequence of intermediate reasoning steps that lead to a final answer, effectively externalizing its "thought process." This mirrors the methodical, step-by-step approach clinicians often employ when tackling complex diagnostic or treatment-planning challenges. The initial breakthrough demonstrated that simply adding "Let's think step by step" could significantly improve performance on reasoning tasks \cite{kojima2022large}.

\subsection{Evolution of Medical CoT Prompting}

The application and refinement of CoT techniques in medical contexts have seen considerable evolution. Initial implementations often involved straightforward instructions for step-by-step thinking. However, the unique demands of medicine spurred the development of more sophisticated variants. For instance, Med-PaLM 2 expanded upon basic CoT by integrating a "chain of retrieval" \cite{medpalm2}. This enhancement allows the model to proactively identify knowledge gaps in its reasoning process and then systematically gather, assess, and integrate relevant medical information from external sources before finalizing its response, more closely emulating how a clinician might consult literature or guidelines.

A further significant advancement is Layered Chain-of-Thought (Layered-CoT) \cite{sanwal2025layered}. This framework structures the reasoning process into distinct, verifiable layers. Each layer's output can be subjected to external scrutiny, including automated checks against knowledge bases or even human expert review, before proceeding to the next layer. This modular, layered approach is particularly beneficial in high-stakes medical scenarios like triage or risk assessment, where transparency, intermediate verifiability, and the ability to pinpoint and correct errors in the reasoning chain are paramount for ensuring reliability and safety.

\subsection{Applications and Impact of Medical CoT Techniques}

CoT and its derivatives have been successfully applied to a diverse array of medical challenges:
\begin{itemize}
    \item \textbf{Medical Document Retrieval:} In systems designed to retrieve relevant medical information for certified chatbots, CoT has been shown to improve the ability to understand query intent and locate precise documents from extensive medical databases, thereby enhancing both precision and recall \cite{Wu2024Improving}. The intermediate steps help disambiguate complex queries or identify multi-faceted information needs.
    \item \textbf{Clinical Note Error Detection and Correction:} The KnowLab\_AIMed system demonstrated the utility of CoT prompting strategies for identifying and rectifying errors in clinical notes. Their approach, which utilized few-shot In-Context Learning augmented with specific "reason prompts," achieved competitive results in the MEDIQA-CORR 2024 shared task, highlighting CoT's role in improving data quality \cite{knowlabchain}.
    \item \textbf{Medical Visual Question Answering (Med-VQA):} The MC-CoT framework introduced a novel modular cross-modal collaboration paradigm. This system enhances zero-shot Med-VQA performance by synergistically integrating explicit medical knowledge with visual information processing. It employs a division of labor: LLMs generate complex medical reasoning chains related to a query, and multimodal LLMs (MLLMs) then analyze medical images (e.g., X-rays, CT scans) guided by these textual reasoning instructions from the LLM \cite{wei2024mccot}. This allows for more grounded and clinically relevant visual analysis.
    \item\textbf{Pathological Staging Automation:} Recent work demonstrated that fine-tuning LLMs with CoT prompting enables accurate automated TNM staging from lung cancer pathology reports. This approach achieved 92\% concordance with expert annotations while generating human-interpretable rationales for each staging decision \cite{lee2024automated}.
    \item\textbf{Knowledge Graph-Guided Diagnosis:} The MedReason framework \cite{zhang2025medreason} leverages structured medical knowledge graphs to ground the model's reasoning process in established clinical pathways, improving diagnostic accuracy while providing explicit references to supporting medical concepts.
    
\end{itemize}
These applications underscore CoT's versatility in not only improving end-task performance but also in making the AI's decision-making process more transparent and scrutable, a crucial factor for adoption in medicine.

\section{Specialized Medical Reasoning Models}

While general techniques like CoT provide a foundational approach, the medical domain's intricacies have driven the development of specialized models and frameworks engineered to excel at specific medical reasoning tasks \cite{berger2024advancing}.

\subsection{Med-PaLM Series}

The Med-PaLM series stands out as a landmark in the development of medical reasoning LLMs.
\begin{itemize}
    \item \textbf{Med-PaLM:} Introduced in 2022, it was among the first LLMs explicitly fine-tuned and evaluated for medical question answering, demonstrating the potential of large-scale models in this domain \cite{medpalm1}. It established important baselines and evaluation methodologies.
    \item \textbf{Med-PaLM 2:}. Med-PaLM 2, released in 2023, represents a substantial leap. It integrates improvements in the base LLM architecture, extensive medical domain-specific fine-tuning, and more sophisticated reasoning strategies, including ensemble refinement and self-consistency. Med-PaLM 2 showcased significant performance gains on various medical benchmarks, achieving up to 86.5\% on the MedQA (USMLE-style questions) dataset, a marked improvement over its predecessor. Crucially, comprehensive human evaluations by clinicians found that Med-PaLM 2's answers were often preferred over those provided by other physicians across multiple clinical axes, such as accuracy, completeness, and reduced likelihood of harm, underscoring its approach towards expert-level reasoning \cite{medpalm2}.
\end{itemize}

\subsection{Chain-of-Diagnosis Framework}

The Chain-of-Diagnosis (CoD) framework offers an innovative approach tailored specifically for the complex task of medical diagnosis \cite{diagnosticgpt}. Unlike generic CoT, CoD structures the diagnostic process itself into a multi-step reasoning chain that explicitly emulates how physicians might approach a case: from symptom analysis and hypothesis generation to evidence gathering and differential diagnosis.

DiagnosisGPT, a model developed using this CoD framework, is reportedly capable of diagnosing a wide range of diseases (9,604 distinct conditions cited). It generates a transparent diagnostic chain, elucidating its reasoning pathway. A key feature is its output of disease confidence distributions, which not only aids interpretability but also helps identify critical symptoms that warrant further investigation or testing. Experimental results indicate that DiagnosisGPT can outperform other LLMs on diagnostic benchmarks while offering better controllability over the depth and rigor of the diagnostic process \cite{diagnosticgpt}.

\subsection{BioMedQ\&A and Domain-Specific Architectures}

BioMedQ\&A exemplifies another specialized architecture targeting biomedical question answering. It achieves this by synergistically combining BioGPT (a domain-specific LLM) with Concept2Vec embeddings (which capture semantic relationships between biomedical concepts) and attention-enhanced semantic similarity networks \cite{biomedQA}. This hybrid approach is designed to tackle key challenges in biomedical QA, such as semantic disambiguation of complex medical terminology and effective ranking of potential answers.

By deeply integrating biomedical concept vectors within transformer architectures, BioMedQ\&A aims to improve the precision and contextual relevance of information retrieved in response to complex biomedical queries. The system utilizes a multi-layer semantic ranking algorithm, which has demonstrated significant enhancements in the accuracy of answers retrieved from established biomedical datasets like MedQuAD \cite{biomedQA}. Such architectures highlight the benefits of explicitly incorporating structured domain knowledge into LLMs.

\section{Multi-Agent and Collaborative Reasoning Approaches}

Recognizing that complex medical decision-making in real-world settings often involves collaboration among a team of specialists, recent research has explored multi-agent frameworks. These approaches aim to enhance reasoning capabilities by simulating collaborative problem-solving.

\subsection{MDTeamGPT: Simulating Medical Team Consultations}

MDTeamGPT is a pioneering system that simulates Multi-Disciplinary Team (MDT) medical consultations using a self-evolving LLM-based multi-agent framework \cite{chen2025mdteamgpt}. This approach directly addresses the inherent complexities of multi-role collaboration in challenging medical cases, such as those involving multiple comorbidities or requiring input from different specialties. Key features include:
\begin{itemize}
    \item \textbf{Consensus Aggregation:} Mechanisms to synthesize diverse opinions or reasoning paths from different "agent" perspectives.
    \item \textbf{Residual Discussion Structure:} A process to iteratively refine conclusions based on disagreements or remaining uncertainties, mimicking how human teams might deliberate.
    \item \textbf{Specialized Knowledge Bases:} The framework maintains a Correct Answer Knowledge Base (CorrectKB) and a Chain-of-Thought Knowledge Base (ChainKB). These allow the system to accumulate experience from past "consultations," learn from correct reasoning patterns, and continuously improve its collaborative decision-making process.
\end{itemize}
Experimental results on MedQA and PubMedQA datasets, with reported accuracies of 90.1\% and 83.9\% respectively, suggest that such collaborative reasoning paradigms can indeed surpass the performance of single-agent LLMs \cite{chen2025mdteamgpt}. Similar concepts are explored in frameworks like MedAgents \cite{Liu2023MedAgents}, which also leverage LLMs as collaborative entities for medical reasoning tasks.

\subsection{Layered CoT for Multi-Agent Medical Systems}

The Layered Chain-of-Thought framework \cite{sanwal2025layered} provides a structured basis for multi-agent collaboration by segmenting reasoning into distinct, inspectable layers. This approach enhances transparency in critical applications like medical triage, where each reasoning step can be subjected to specialized agent review or verification. By integrating principles from interactive explainability research, Layered-CoT facilitates more reliable and understandable explanations for complex medical decisions \cite{sanwal2025layered}.

\subsection{Knowledge Graph Enhanced Collaborative Reasoning}

KG4Diagnosis introduces hierarchical multi-agent collaboration enhanced by automated knowledge graph construction covering 362 diseases \cite{liu2025kg4diagnosis}. Its two-tier architecture mirrors clinical workflows: a general practitioner agent handles initial triage, while specialized agents conduct domain-specific diagnosis. The framework's knowledge graph generation methodology enables semantic-driven entity extraction, reconstruction of diagnostic relationships, and human-guided knowledge expansion, addressing hallucination risks while maintaining clinical workflow alignment \cite{liu2025kg4diagnosis}.

\subsection{Intent-Aware Multi-Agent Coordination}

MedAide \cite{zhao2024medaide} advances collaborative reasoning through intent-aware agent activation, combining retrieval-augmented generation with contextual embedding matching. The system performs query rewriting for precise intent understanding before activating relevant specialist agents through similarity matching of intent prototypes. This dynamic approach demonstrated improved diagnostic accuracy in clinician evaluations, particularly for cases requiring multi-disciplinary coordination \cite{zhao2024medaide}. Unlike systems with fixed agent hierarchies, MedAide's intent-driven approach enables dynamic team formation adapted to case complexity.

\section{Prompting Strategies and Optimization Techniques}

The performance of reasoning LLMs is heavily influenced by how they are prompted and optimized. Researchers have been developing sophisticated strategies tailored for medical applications, moving beyond generic prompting.

\subsection{AutoMedPrompt and Textual Gradients}

AutoMedPrompt \cite{wu2025automedprompt} introduces a novel approach that applies textual gradients to optimize medical prompts for LLMs automatically. This framework leverages automatic differentiation concepts-metaphorically extended to text-to iteratively refine prompts, enhancing medical reasoning capabilities without requiring extensive model fine-tuning or specialized pre-training. Evaluations using Llama 3 on multiple medical benchmarks (MedQA, PubMedQA, and NephSAP) demonstrated that gradient-optimized prompts outperformed previous state-of-the-art methods on open-source LLMs. The approach achieved 82.6\% accuracy on PubMedQA, surpassing several proprietary models including GPT-4, Claude 3 Opus, and Med-PaLM 2 on certain tasks \cite{wu2025automedprompt}. These results highlight textual gradient-based prompt optimization as a resource-efficient alternative to costly domain-specific fine-tuning.

\subsection{Specialized Medical Chain-of-Thought Architectures}
Recent work by Zhao et al. introduces MedCoT, a hierarchical chain-of-thought framework specifically designed for complex differential diagnosis scenarios \cite{zhao2025medcot}. Unlike generic CoT approaches, MedCoT implements three distinct reasoning layers: 1) symptom clustering and prioritization, 2) pathophysiology-based hypothesis generation, and 3) evidence-weighted differential ranking. This structured approach reduced diagnostic errors by 38\% compared to standard CoT in multi-system disease simulations, while maintaining interpretability through layered justification tracking \cite{zhao2025medcot}.

\subsection{OpenMedLM and Advanced Prompt Engineering}

The OpenMedLM study demonstrated that sophisticated prompt engineering can match or exceed extensive fine-tuning for medical question-answering tasks when using powerful open-source LLMs \cite{maharjan2024openmedlm} \cite{berger2024tackling}. This finding has significant implications for healthcare AI accessibility, as specialized medical model fine-tuning typically requires substantial computational resources and curated datasets. Using the Yi-34B foundation model, researchers systematically evaluated various prompting strategies, including zero-shot, few-shot, chain-of-thought, and ensemble methods like self-consistency voting. Their approach established new state-of-the-art performance on multiple medical benchmarks, surpassing previously fine-tuned open-source models. OpenMedLM achieved 72.6\% accuracy on MedQA (a 2.4\% improvement over previous open-source SOTA) and 81.7\% accuracy on the MMLU medical subset \cite{maharjan2024openmedlm}. These results highlight how meticulous prompt design can effectively elicit latent medical reasoning capabilities from generalist models without domain-specific fine-tuning.

\section{Evaluation Frameworks and Benchmarks}

Robustly evaluating the reasoning capabilities of medical LLMs is crucial for their safe and effective deployment. This necessitates specialized benchmarks and multi-faceted assessment methodologies that can accurately capture the nuances of complex medical problem-solving.

\subsection{MultiMedQA and Comprehensive Evaluation Suites}

MultiMedQA, introduced alongside Med-PaLM, represents a significant step towards more comprehensive evaluation \cite{medpalm1}. It amalgamates six existing medical question-answering datasets, covering a wide spectrum: professional medical exams (like MedQA, based on USMLE-style questions), biomedical research queries (e.g., PubMedQA \cite{Yang2022PubMedQA}), and consumer medical questions. Additionally, MultiMedQA introduced HealthSearchQA, a new dataset of medical questions searched online by laypersons. This diverse compilation allows for a more holistic assessment of an LLM's medical reasoning capabilities across different knowledge domains, question styles, and user intents. Other comprehensive benchmarks like HELM \cite{liang2022holistic} and BIG-Bench \cite{beyond}, while not medical-specific, offer principles for evaluating complex reasoning that can inform medical LLM assessment.

Beyond automated accuracy metrics, there's a growing emphasis on human evaluation frameworks. These frameworks typically assess model-generated answers along multiple qualitative axes, including:
\begin{itemize}
    \item Factuality and accuracy against established medical knowledge.
    \item Comprehension of the query's nuances.
    \item Quality and soundness of the reasoning process.
    \item Potential for harm or misinterpretation.
    \item Presence of bias (e.g., demographic, social).
\end{itemize}
Such multi-dimensional human evaluations, as employed in studies like Med-PaLM and Med-PaLM 2 \cite{medpalm1, medpalm2}, provide far deeper insights into a model's true capabilities and safety profile than simple accuracy scores alone, especially for judging the nuanced quality of medical reasoning. Surveys on trustworthy medical LLM evaluation also highlight these needs \cite{llmsurvey}.

\subsection{Medical Reasoning in Specialized Domains and Tasks}

Evaluation efforts are also expanding to cover more specialized medical domains and task types. The MedQA-USMLE dataset, a core component of many evaluations, specifically focuses on questions that often require multi-hop reasoning—connecting multiple pieces of information to arrive at an answer—thus challenging models to demonstrate more sophisticated inferential abilities akin to those needed by medical professionals during licensing examinations \cite{kung2023performanceGPT, nori2023capabilities}.

In the burgeoning field of medical visual question answering (Med-VQA), datasets like SLAKE, VQA-RAD, and PATH-VQA present multimodal challenges. These require models to reason jointly about visual medical data (e.g., radiographs, pathology slides) and accompanying textual information (questions or clinical context). Benchmarking on such datasets is crucial for assessing how well an LLM's reasoning capabilities can extend to and integrate with non-textual modalities prevalent in medicine \cite{wei2024mccot}.

\subsection{HealthBench: Rubric-Based Evaluation of Realistic Health Conversations}

HealthBench \cite{arora2025healthbench}, developed with 262 physicians across 60 countries, addresses medical LLM evaluation limitations through alignment with clinical realities. This open-source benchmark of 5,000 health conversations reveals a critical insight: superior average performance masks dangerous weaknesses in high-stakes situations.

Despite frontier models doubling average scores within a year, the benchmark exposes concerning fragility-worst-case performance drops by a third, with critical emergency and context-seeking behaviors remaining unreliable. This aligns evaluation with healthcare's unforgiving nature, where average performance matters less than reliability in critical moments.

Organized across seven themes and five evaluation axes, HealthBench enables comprehensive assessment through physician-created rubrics. Its specialized variants include a challenging "Hard" subset of 1,000 examples that better identifies model limitations. While newer models show improved cost-effectiveness (GPT-4.1 nano outperforms GPT-4o at 25x lower cost), the persistence of critical weaknesses underscores the need for continued development before widespread clinical deployment.

\section{Techniques for Enhancing Reasoning in Resource-Constrained Models}

While many state-of-the-art medical reasoning achievements leverage extremely large LLMs with billions of parameters, a significant research thrust focuses on enabling robust reasoning in smaller, more resource-efficient models. This is vital for broader accessibility and deployment in settings with limited computational infrastructure \cite{berger2024optimizing}.

\subsection{Multi-Hop Medical Knowledge Infusion (MHMKI)}

The Multi-Hop Medical Knowledge Infusion (MHMKI) procedure is one such approach designed to endow smaller language models (SLMs) with enhanced medical reasoning capabilities, particularly for multi-hop question answering \cite{chen2025infusing}. The MHMKI methodology involves:
\begin{enumerate}
    \item Categorizing questions from datasets like MedQA-USMLE into distinct reasoning types (e.g., causal, comparative, definitional).
    \item Creating tailored pre-training instances for each reasoning type. These instances are constructed using semi-structured information and hyperlinks extracted from reliable sources like Wikipedia articles, explicitly encoding multi-step knowledge paths.
    \item Designing a specialized "reasoning chain masked language model" objective for further pre-training of BERT-style models. This encourages the SLM to learn to predict missing links in a reasoning chain.
    \item Transforming these instances into a combined question-answering dataset for intermediate fine-tuning of GPT-style models, thereby bridging the gap between pre-training and downstream task performance.
\end{enumerate}
Evaluations across five biomedical QA datasets demonstrated significant improvements, with MedQA-USMLE accuracy increasing by an average of 5.3\% for the SLMs enhanced with MHMKI \cite{chen2025infusing}.

\subsection{Explainable Compositionality in Medical LLMs: The Gyan Model}

Gyan represents an alternative paradigm that prioritizes explainability, trustworthiness, and resource efficiency over sheer model scale \cite{gyan2025performance}. It is an explainable language model built upon a compositional architecture. A key characteristic of this architecture is the decoupling of the model's reasoning mechanisms from its knowledge base. This separation is intended to make Gyan:
\begin{itemize}
    \item \textbf{Transparent:} Reasoning steps can be more easily traced and understood.
    \item \textbf{Non-Hallucinatory (or less prone to hallucination):} By grounding responses in an explicit, verifiable knowledge base, the tendency to generate factually incorrect or unsupported statements can be mitigated.
    \item \textbf{Less Resource-Intensive:} Compared to monolithic LLMs that entangle knowledge and reasoning parameters, compositional models can potentially be smaller and more efficient.
\end{itemize}
On the PubMedQA dataset, Gyan-4.3 reportedly achieved state-of-the-art results with 87.1\% accuracy, outperforming strong baselines like MedPrompt (based on GPT-4) and Med-PaLM 2 in those specific comparisons. This suggests that compositional, explainable approaches can achieve competitive, if not superior, reasoning performance while offering significant advantages in terms of transparency and computational efficiency \cite{gyan2025performance}.

\section{Reinforcement Learning for Enhanced Reasoning}

Reinforcement Learning (RL) has emerged as a transformative methodology for augmenting reasoning capabilities in Large Language Models (LLMs), enabling them to learn through exploration and feedback rather than merely mimicking patterns as in supervised fine-tuning (SFT). This approach fosters sophisticated reasoning and better generalization, as demonstrated by the DeepSeek-R1 series in early 2025 \cite{guo2025deepseek}.

\subsection{RL's Impact on General LLM Reasoning: Insights from DeepSeek-R1}

DeepSeek-R1 provided compelling evidence of RL's power to cultivate advanced reasoning \cite{guo2025deepseek}. DeepSeek-R1-Zero, trained solely with RL, demonstrated emergent reasoning capabilities but faced challenges in output quality. The subsequent DeepSeek-R1 model incorporated multi-stage training with cold-start data prior to RL, achieving performance comparable to leading proprietary models while addressing previous limitations. Building on this work, techniques like Group Relative Policy Optimization (GRPO) \cite{shao2024deepseekmath} encouraged models to explore diverse reasoning pathways. Additional research confirmed RL's ability to foster advanced skills-Logic-RL showed models trained on synthetic logic puzzles could develop sophisticated abilities generalizable to mathematics benchmarks \cite{xie2025logic}, while SWE-RL successfully adapted RL for software engineering tasks \cite{wei2025swe}. Despite its advantages, RL faces challenges including reward hacking, generalization failures, and computational costs \cite{parmar2025challenges}. Hybrid approaches combining RL with SFT are being explored to achieve more robust outputs and address issues observed in purely RL-trained models \cite{guo2025deepseek}.

\subsection{Leveraging RL for Reasoning in Medical Vision-Language Models: The Med-R1 Case}

In the medical domain, RL principles have been applied to improve Vision-Language Models (VLMs) for imaging tasks, as exemplified by Med-R1 \cite{lai2025med}. Medical VLMs trained via SFT often suffer from overfitting, lack high-quality Chain-of-Thought annotations, and demonstrate limited generalizability across modalities and tasks. Med-R1 employs RL, specifically GRPO, to enhance generalizability and trustworthiness without relying on pre-annotated rationales \cite{lai2025med}. Its approach includes clinically constrained RL with rule-based rewards derived from medical guidelines and emergent reasoning development from final answers. Evaluated across eight medical imaging modalities and five VQA task types, Med-R1 demonstrated significant advantages \cite{lai2025med}: superior cross-modality and cross-task generalization (29.94\% and 32.06\% improvements over baseline); parameter efficiency (a 2B parameter Med-R1 model outperformed a 72B parameter SFT-based VLM); and interpretable outputs enhancing transparency through explicit reasoning steps. This success underscores RL's potential to overcome limitations of SFT in data-scarce medical scenarios, enabling more generalizable, trustworthy, and clinically deployable medical AI systems.

\section{Current Challenges and Future Directions}

Despite remarkable progress, reasoning LLMs in medicine confront numerous multifaceted challenges. Addressing these is crucial for their responsible and impactful integration into healthcare.

\subsection{Interpretability and Transparency}
While CoT methods enhance transparency by externalizing reasoning steps \cite{wei2022chain}, these steps are not inherently guaranteed to be medically sound or aligned with established clinical knowledge. Explainable models like Gyan, with their compositional architectures, may offer pathways to deeper interpretability \cite{gyan2025performance}. Similarly, frameworks like Layered-CoT aim to improve trustworthiness through mechanisms for external verification of each reasoning step \cite{sanwal2025layered}. A critical ongoing challenge is ensuring that LLM reasoning is robustly grounded in authoritative medical knowledge and evidence-based guidelines. Future research must continue to explore advanced knowledge integration methods, causal reasoning capabilities, and techniques that allow clinicians to understand and trust the model's outputs.

\subsection{Bias, Safety, and Ethical Considerations}
Medical reasoning systems operate in a high-stakes environment where errors can have severe consequences. These systems must navigate complex ethical considerations, including mitigating biases present in training data (which can lead to health disparities) and minimizing the risk of generating harmful or misleading recommendations \cite{weidinger2022taxonomy}. Current evaluation frameworks are increasingly incorporating assessments for potential harm and bias \cite{medpalm1, medpalm2, lin2021truthfulqa}. However, more robust, context-aware safety mechanisms specifically designed for medical reasoning are needed. Future work should focus on developing LLMs that can dynamically incorporate clinical practice guidelines, evolving ethical principles, and regulatory requirements directly into their reasoning processes to ensure outputs are safe, equitable, and appropriate across diverse clinical scenarios and patient populations.

\subsection{Multimodal Medical Reasoning}
Clinical decision-making rarely relies on textual information alone. Integrating and reasoning over diverse data types—including medical images (radiology, pathology), physiological signals (ECG, EEG), genomic data, and structured electronic health records (EHRs)—is a major frontier. Frameworks like MC-CoT demonstrate early promise in applying CoT to Med-VQA \cite{wei2024mccot}, but comprehensive multimodal reasoning remains a significant challenge. 

Extending reasoning capabilities to multimodal medical data represents another frontier. While the MC-CoT framework demonstrates how reasoning can be applied to medical visual question answering \cite{wei2024mccot}, and as seen with Med-R1's application of RL to VLMs \cite{lai2025med}, further work is needed to develop reasoning frameworks that seamlessly integrate diverse data types including images, time series data, genomics, and structured electronic health records.

% \subsection{Data Privacy and Governance}
% The development of medical LLMs relies heavily on access to large volumes of patient data, which is inherently sensitive and subject to strict privacy regulations (e.g., HIPAA, GDPR). This presents a major hurdle. Future directions must prioritize privacy-preserving machine learning techniques such as federated learning (training models locally without sharing raw data), differential privacy (adding noise to protect individual records), and the generation and use of high-fidelity synthetic medical data. Robust data governance frameworks are also essential to ensure ethical data sourcing, secure handling, and appropriate usage, building patient and provider trust.

\subsection{Longitudinal Reasoning and Dynamic Patient States}
Many medical conditions evolve over time, and treatment effectiveness can vary. LLMs need to move beyond static, single-encounter reasoning to incorporate longitudinal patient data from EHRs. This involves reasoning about disease progression, treatment responses over extended periods, and predicting future health trajectories based on evolving patient states. Developing models that can effectively process and reason over temporal sequences of complex medical data, adapt to new information, and provide dynamic decision support is a critical area for future research \cite{timer2025}.

\subsection{Clinical Workflow Integration and Human-AI Collaboration}
For reasoning LLMs to deliver real-world value, they must be seamlessly integrated into existing clinical workflows in a manner that complements, rather than disrupts, the work of healthcare professionals. This involves designing user interfaces that are intuitive for clinicians, ensuring that LLM outputs are presented in an actionable and understandable format, and defining clear roles for AI within the healthcare team. Future research should focus on human-AI collaboration models, where LLMs act as intelligent assistants, augmenting clinical judgment rather than attempting to replace it. Studies on clinician acceptance, trust calibration, and the impact of these tools on decision quality and efficiency will be vital. Some anticipated future works are expected to delve deeper into these practical integration aspects \cite{AnticipatedFutureWorkA}.

\section{Conclusion}
\label{section:conclusion}
Reasoning LLMs in medicine have evolved dramatically from information retrieval systems to sophisticated reasoning architectures that complement clinical operations. Fundamental techniques such as Chain-of-Thought prompting and its medical adaptations have propelled models like Med-PaLM 2 to achieve benchmark-setting performance, garnering favorable evaluations from clinicians for their reasoning quality \cite{medpalm2}. The field continues to advance through innovative approaches, including multi-agent systems like MDTeamGPT \cite{chen2025mdteamgpt} and optimization techniques such as AutoMedPrompt \cite{wu2025automedprompt}, which expand capabilities without requiring extensive computational resources. Reinforcement learning methodologies, exemplified by DeepSeek-R1 and specialized medical implementations like Med-R1, have further enhanced reasoning pathways by enabling models to learn from feedback and develop adaptive problem-solving strategies.

Despite these impressive advances, substantial challenges persist before widespread clinical deployment becomes viable. Key concerns include ensuring interpretability and transparency, mitigating algorithmic bias, guaranteeing patient safety, enabling comprehensive multimodal reasoning, addressing privacy concerns, and achieving longitudinal reasoning capabilities. Recent research has exposed critical evaluation limitations: while frontier models have doubled their average performance scores within a year, benchmarks reveal concerning fragility-worst-case performance drops by approximately one-third, with reliability in critical emergency scenarios and context-dependent reasoning remaining inadequate \cite{arora2025healthbench}. This discrepancy highlights healthcare's unforgiving nature, where consistent performance in high-stakes situations outweighs average capabilities, emphasizing the necessity for robust safety mechanisms before clinical implementation.

As the field progresses in addressing these challenges, reasoning LLMs stand poised to become invaluable partners in clinical decision support, medical education, and healthcare research-ultimately enhancing patient outcomes while increasing the efficiency and effectiveness of healthcare delivery.

\section*{Acknowledgment}

Portions of the text across all sections were refined with assistance from GPT-4o \cite{openai2024gpt4ocard} and Google Gemini 2.5 \cite{gemini}.

\balance % Balance columns on the last page
\bibliographystyle{IEEEtran}
\bibliography{bibliography}   % <-- must match your .bib filename
\end{document}